\documentclass[10pt,twocolumn,letterpaper]{article}

\usepackage{cvpr}

\usepackage{amsfonts}
\usepackage{textcomp}
\usepackage{multirow}
\usepackage{algorithmic}
\usepackage[ruled,linesnumbered,noline]{algorithm2e}
\SetNlSty{}{}{:}

\definecolor{cvprblue}{rgb}{0.21,0.49,0.74}
\usepackage[pagebackref,breaklinks,colorlinks,allcolors=cvprblue]{hyperref}

\def\confName{CVPR}
\def\confYear{2026}

\title{C$^2$GR: Coupled Comprehensive Generative Replay for a Continually Learnable Universal Segmentation Model}

\author{
Wei Li$^{1}$ \quad
Jingyang Zhang$^{2}$\textsuperscript{*} \quad
Guoan Wang$^{3}$ \quad
Junzhi Ning$^{4}$ \quad
Yang Chen$^{2}$\\
Guang Yang$^{5}$ \quad
Lixu Gu$^{1}$\textsuperscript{*}\\
$^{1}$Shanghai Jiao Tong University \quad
$^{2}$Southeast University \quad
$^{3}$Stevens Institute of Technology\\
$^{4}$Shanghai Artificial Intelligence Laboratory \quad
$^{5}$Imperial College London\\
{\tt\small liwei2022@sjtu.edu.cn, j.y.zhang@seu.edu.cn, gulixu@sjtu.edu.cn}\\
}

\begin{document}
\maketitle
% !TEX root = ../main.tex
\begin{abstract}

Universal segmentation models exhibit significant potential for diverse tasks involving different imaging modalities and segmentation objectives.
Task-Incremental Learning provides a privacy-preserving approach to continually evolve a universal model on tasks from sequentially-arriving medical departments.
However, training the model solely on the incoming task induces forgetting on past tasks, since consecutive tasks exhibit concurrent shifts in image appearance and segmentation objective.
To address this problem, we propose a novel Coupled Comprehensive Generative Replay (C$^2$GR) framework that simultaneously synthesizes image-mask pairs of previous tasks to mitigate forgetting under concurrent appearance and objective shifts.
This requires preserving image-mask correspondence for structure-realistic generation and bridging asynchronous optimization of the generator and segmentor for segmentation-oriented generation.
Specifically, we propose a Bayesian Joint Diffusion (BJD) method that formulates the correspondence as conditional distributions optimized via conditional denoising.
Furthermore, we develop a Relation-aware Unified Prompt Synchronization (RUPS) scheme to simultaneously modulate the generator and segmentor via a shared task-relation-aware prompt for synchronizing their optimization.
Experiments on 20 tasks spanning diverse modalities and objectives demonstrate that C$^2$GR exhibits only a 2.44\% drop in overall performance compared to joint training with all task data, effectively alleviating forgetting from the concurrent shifts.
Our code will be made publicly available at \url{https://github.com/mar-cry/C2GR}.

\end{abstract}

\noindent\textbf{Keywords:} Incremental Learning, Generative Replay, Joint Diffusion Model, Prompt Synchronization.

\par\vfill
\noindent\rule{0.42\columnwidth}{0.4pt}\par
{\footnotesize\textsuperscript{*}Corresponding authors: Jingyang Zhang and Lixu Gu.\par}
% !TEX root = ../main.tex

\section{Introduction}
\label{sec:intro}

\begin{figure}[!t]
\centering
\includegraphics[width=\columnwidth]{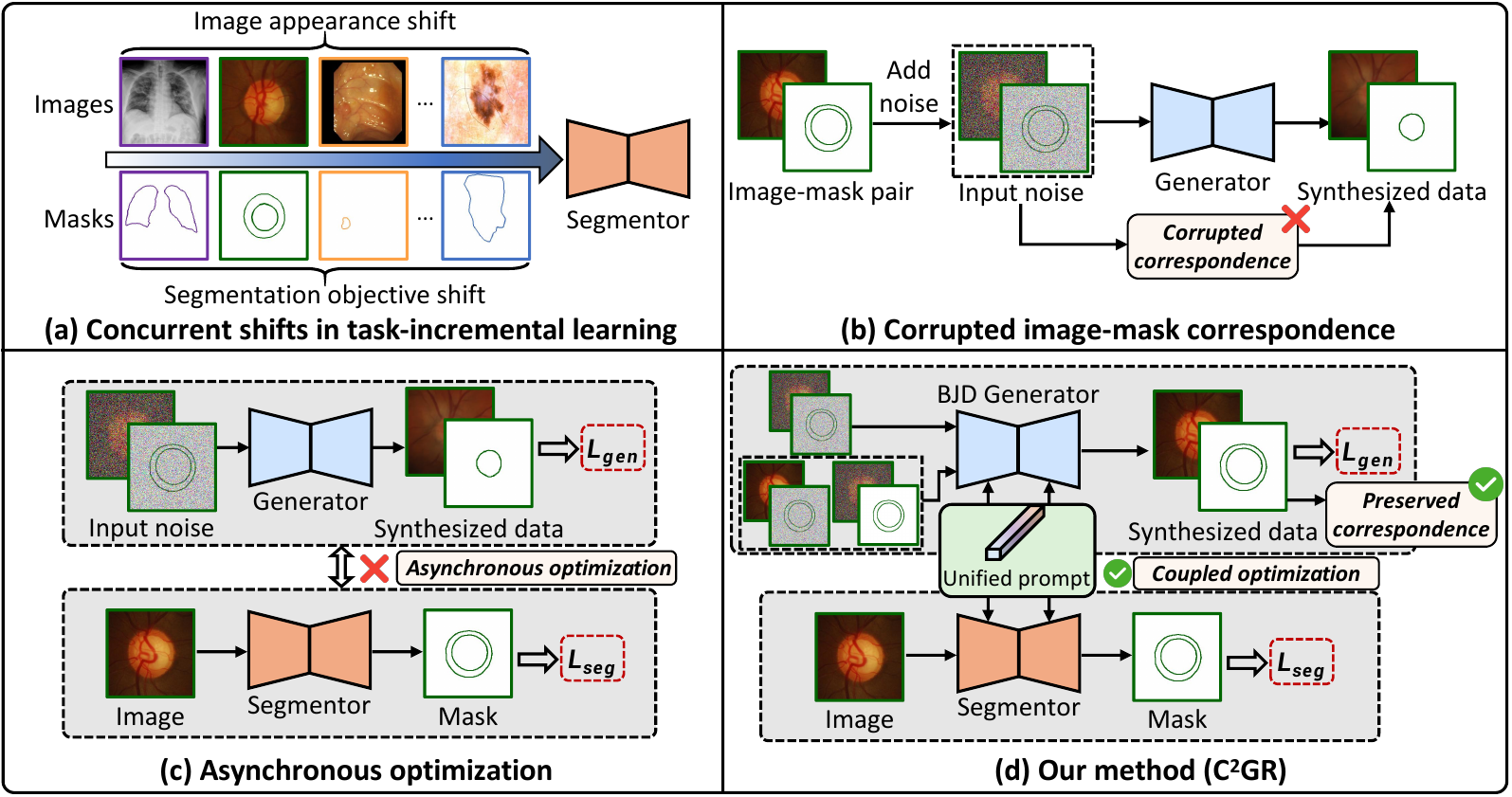}
\caption{(a) Task-incremental learning introduces concurrent shifts in image appearance and segmentation objective. Directly employing diffusion models to generate image-mask pairs of past tasks for overcoming the concurrent shifts presents two limitations: (b) Corrupted image-mask correspondence caused by noise naively added to both image and mask, and (c) Asynchronous optimization between the generator and segmentor introducing potential generation error that compromises knowledge retention of the segmentor. (d) Our method preserves image-mask correspondence via the proposed BJD method and couples optimization of the generator and segmentor through a unified prompt.}
\label{fig:motivation}
\end{figure}

In medical image segmentation, increasing demand has emerged for a universal model capable of addressing diverse tasks across varying modalities, anatomical regions, and segmentation objectives \cite{ma2024segment,liu2023clip,zhang2025generalist}.
Existing universal models demonstrate promising multi-task performance \cite{zhang2025generalist} and could relieve clinical experts from manually selecting the optimal model for each task \cite{ma2024segment}.
Such success relies on a static learning paradigm \cite{likangsrdsfw}, which first aggregates all data from diverse clinical departments and then conducts model training.
However, this paradigm would be impractical in clinical scenarios, as it neglects stringent data-sharing policies \cite{price2019privacy} and the necessity of updating the model as new segmentation tasks emerge \cite{sammedmoe}.

Task-Incremental Learning (TIL) \cite{cgr} has emerged as a privacy-preserving and incrementally expandable paradigm that enables the universal model to learn from sequentially arriving tasks without access to previous task data.
However, substantial task heterogeneity arises under TIL, since successive tasks may originate from diverse anatomical regions with varying imaging devices and segmentation objectives \cite{ma2024segment}, as illustrated in Fig.~\ref{fig:motivation}(a).
Due to such \emph{concurrent shifts in image appearance and segmentation objective}, training the universal model solely on the incoming task could induce severe performance degradation on previously learned tasks.

% 这一段引用文献的时候可以加上这些related works需要添加一些近期的比如prompt-based工作
Nevertheless, although appearance and objective shifts co-occur in TIL, existing studies treat them as isolated problems and address each through specialized paradigms.
Specifically, Class-Incremental Learning (CIL) mitigates forgetting under objective shift within a confined anatomical region exhibiting no or marginal appearance shift \cite{plop,meil,conuseg,promptcl}.
CIL methods typically employ knowledge distillation \cite{plop} and semantic prototype transfer on model features \cite{conuseg,meil} or lightweight prompts \cite{promptcl} to preserve objective-specific knowledge.
However, considering the TIL scenario where successive objectives originate from different anatomical regions with substantial appearance discrepancies, the appearance shift would undermine the reliability of semantic representations during distillation or transfer \cite{ted}.
Additionally, Domain-Incremental Learning (DIL) addresses forgetting under appearance shift for a fixed objective via style regularization \cite{kirkpatrick2017overcoming,s3r,ted} and appearance-focused generative replay \cite{likangsrdsfw,gr}.
Yet, in TIL where appearance shift may coexist with objective changes, style-oriented regularization would struggle to capture feature variations induced by the changed objective \cite{dc2t}. 
Meanwhile, appearance-focused generative replay would be dominated by the current task objective due to synthesizing only images and adopting error-prone masks biased toward the current task \cite{likangsrdsfw}.
Overall, although existing methods address objective or appearance shift in isolation, they cannot simultaneously tackle both issues inherent in TIL.

To alleviate forgetting caused by concurrent appearance and objective shifts, inspired by data-centric rehearsal \cite{likangsrdsfw,gr,smglearning}, our insight involves simultaneously synthesizing images and corresponding masks for previous tasks via diffusion models with advanced visual generation capability in the medical field \cite{medsegfactory}.
However, this strategy presents two key challenges.
First, as illustrated in Fig.~\ref{fig:motivation}(b), directly generating image-mask pairs via diffusion models would result in \emph{corrupted image-mask correspondence} when noise is added to both the image and mask simultaneously during the forward diffusion process \cite{medsegfactory}.
The corruption originates from the noise schedule degenerating the joint distribution of image and mask into a product of their marginal distributions.
Therefore, it is desired to analyze the composition of the joint distribution via Bayes' theorem and recover it from the marginal distributions to consolidate the image-mask correspondence.
Second, as shown in Fig.~\ref{fig:motivation}(c), existing generative replay methods widely adopt \emph{asynchronous optimization between the generator and segmentor} \cite{gr}, which optimizes the two models independently.
Directly employing such independent optimization strategy may not guarantee a segmentation-oriented generation process, thereby introducing generation error that compromises the knowledge retention of the segmentor.
To address this problem, our insight is to leverage the task-relation prior \cite{taskprior} shared between the generator and segmentor to bridge their optimization processes.
This is motivated by the observation that the shared task-relation prior could be reinforced for segmentation during segmentor optimization \cite{taskprior} and would simultaneously interact with the generator during generator optimization to promote segmentation-oriented generation \cite{zhang2025generative}.

In this paper, we propose \emph{to our knowledge the first} TIL framework for training a universal segmentation model capable of handling diverse tasks.
Our framework, called Coupled Comprehensive Generative Replay (C$^2$GR), mitigates forgetting under concurrent appearance and objective shifts by simultaneously synthesizing image-mask pairs of previous tasks.
To generate effective image-mask pairs for preserving segmentor knowledge, as shown in Fig.~\ref{fig:motivation}(d), it is essential to ensure structure-realistic synthesis by preserving the image-mask correspondence, and segmentation-oriented generation by coupling the asynchronous optimization of the generator and segmentor.
Specifically, to preserve the image-mask correspondence, we propose a Bayesian Joint Diffusion (BJD) method that formulates the image-mask correspondence as conditional distributions based on Bayes' theorem and optimizes the conditional distributions through conditional denoising.
To couple the asynchronous optimization, we introduce a Relation-aware Unified Prompt Synchronization (RUPS) mechanism that formulates a task-relation-aware prompt and injects it into both the generator and segmentor for simultaneously updating both models.
The contributions are as follows:
\begin{itemize}
\item We study the forgetting issue of building a universal model via task-incremental learning under concurrent shifts in image appearance and segmentation objective, and propose a novel Coupled Comprehensive Generative Replay (C$^2$GR) framework.

\item We propose a Bayesian Joint Diffusion (BJD) method to preserve the essential image-mask correspondence when generating image-mask pairs for previous tasks.

\item We introduce a Relation-aware Unified Prompt Synchronization (RUPS) mechanism to couple the asynchronous optimization of the generator and the segmentor.

\item We conduct extensive experiments on 20 segmentation tasks spanning 8 imaging modalities and 9 anatomical regions to validate the superiority of the proposed method over existing DIL and CIL methods.
\end{itemize}

A preliminary version of this work was presented in \cite{cgr}. This journal version expands upon it by (1) proposing Coupled Comprehensive Generative Replay for training a universal segmentation model via task-incremental learning; (2) introducing a Relation-aware Unified Prompt Synchronization mechanism to bridge the optimization of the generator and segmentor; (3) verifying the effectiveness of our method on 20 tasks spanning 8 imaging modalities and 9 anatomical regions; and (4) conducting more comprehensive comparisons with DIL and CIL methods and more extensive ablation studies.

% !TEX root = ../main.tex
\section{Related Work}
\label{sec:Related Work}

\subsection{Domain-Incremental Learning}

Domain-incremental learning (DIL) aims to mitigate forgetting under appearance shift when a model learns from sequentially arriving sites (i.e., domains), where appearance variations exist across different domains with fixed segmentation objectives.
Existing DIL methods mainly fall into two categories: 1) style regularization-based methods, which preserve appearance-related knowledge by penalizing parameters important to earlier domains \cite{kirkpatrick2017overcoming,s3r}, or by distillation focusing on appearance-related knowledge \cite{ted}; and 2) appearance replay-based methods, which either maintain a replay buffer storing raw data from previous domains \cite{smglearning}, or utilize generative replay to synthesize images of previous tasks from random noise \cite{gr} or from segmentation masks of the current task \cite{likangsrdsfw}.
However, these methods primarily retain appearance-related knowledge from previous domains, while overlooking degradation caused by changes in segmentation objectives.

\subsection{Class-Incremental Learning}

Class-incremental learning (CIL) focuses on learning from sequentially arriving tasks with different segmentation objectives without forgetting, assuming that the appearance shift across tasks is slight or even negligible.
Existing CIL methods can be divided into three categories: 1) objective knowledge distillation \cite{plop,ye2025medseqft}, which transfers previous model features related to the segmentation objective to the current model; 2) semantic prototype transfer \cite{meil,conuseg}, which extracts semantic prototypes associated with segmentation objectives and preserves the relation between objective-specific prototypes; and 3) architecture-based methods \cite{sammedmoe}, which allocate task-specific parameters to preserve objective-related knowledge of each task.
However, existing CIL methods ignore the appearance shift present across different segmentation objectives. Such appearance shift could corrupt extraction of objective-related knowledge, thereby exacerbating forgetting for previous tasks.

\subsection{Incremental Learning for Universal Model}

Few studies have investigated incremental learning (IL) for training a universal model, where appearance shift and segmentation-objective shift occur concurrently.
Existing works employ naive architecture-based strategies (e.g., SAM-Med3D-MoE \cite{sammedmoe} and Continual SAM \cite{wang2026continual}) or regularization-based strategies (e.g., MedSeqFT \cite{ye2025medseqft}) from DIL or CIL to alleviate forgetting under concurrent shifts.
The architecture-based methods typically apply parameter-efficient, task-specific adapters \cite{sammedmoe} to modulate the frozen pre-trained backbone, while the regularization-based strategies penalize knowledge sensitive to task changes from being updated.
However, these methods only adapt a pre-trained model to incoming tasks on a small number of tasks, rather than developing a universal model across diverse tasks from scratch.
Directly applying naive DIL or CIL strategies could hinder scalability in long-term task-incremental learning with diverse modalities and objectives \cite{ma2024segment}, especially when training a universal model.
These drawbacks motivate the need for a more reliable IL method to recover both appearance-related and objective-related knowledge for continually training a universal model.

% !TEX root = ../main.tex
\section{Methodology}
\label{sec:method}

\begin{figure*}[!t]
\centerline{\includegraphics[width=\textwidth]{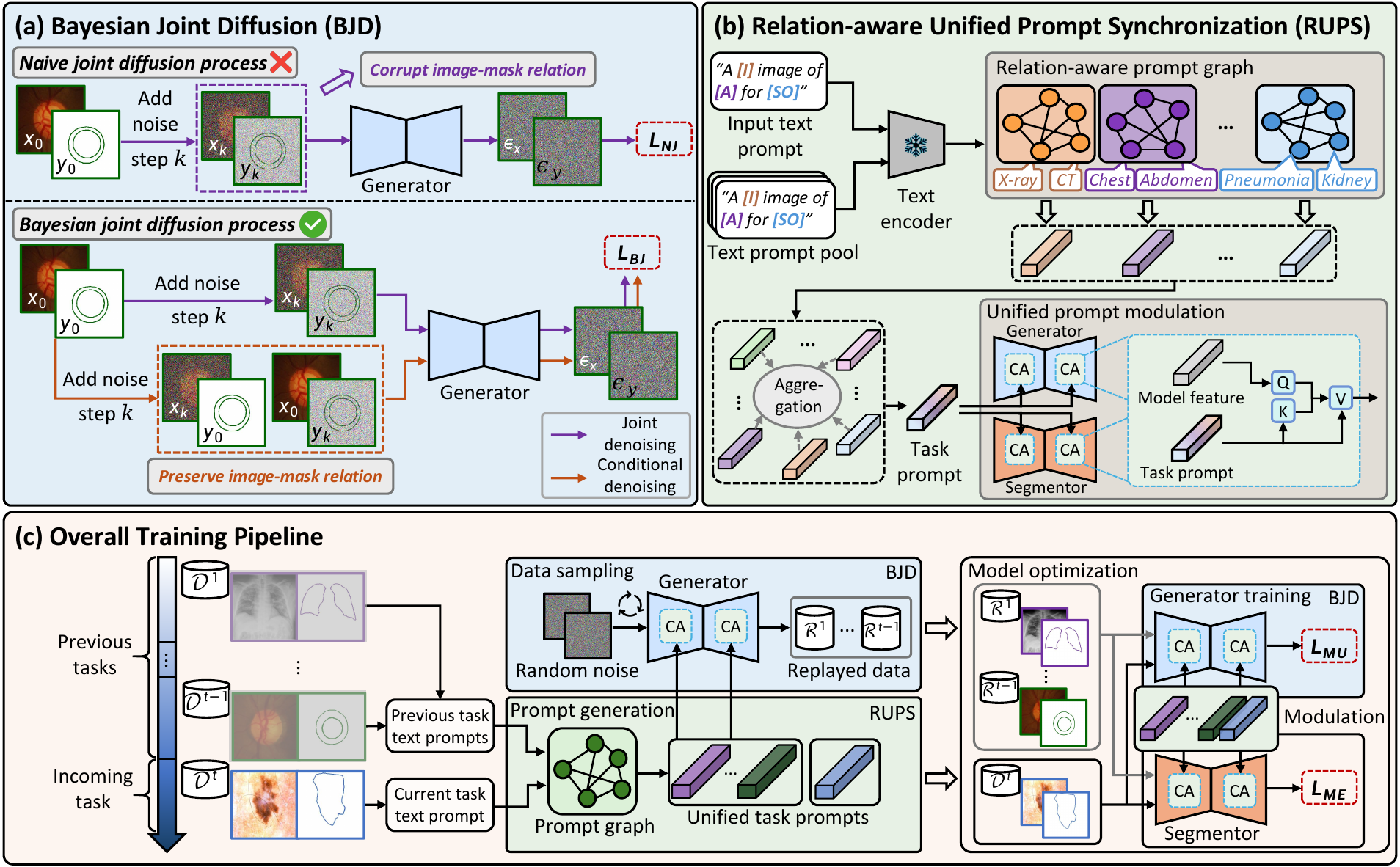}}
\caption{Overview of Coupled Comprehensive Generative Replay (C$^2$GR). C$^2$GR mitigates forgetting by synthesizing image-mask pairs for previous tasks, where the core is to preserve image-mask correspondence and couple the optimization of generator and segmentor. (a) We adopt Bayesian Joint Diffusion (BJD, Sec.~\ref{sec:BIMJD}) to preserve image-mask correspondence by formulating it as conditional distributions optimized via conditional denoising. (b) We design Relation-aware Unified Prompt Synchronization (RUPS, Sec.~\ref{sec:rupg}) that leverages relation-aware prompt graph to produce the task prompt, which is injected into both the segmentor and generator for further synchronizing the optimization of both models. (c) The overall training pipeline upon a new task arrival, consisting of replay data synthesis, prompt generation, and coupled optimization as described in Sec.~\ref{sec:training}.
}
\label{fig:overall}
\end{figure*}

When employing Task-Incremental Learning (TIL) to develop a universal segmentation model $f_{\theta}$, the data stream comprises heterogeneous tasks with concurrent shifts in image appearance and segmentation objective.
At each learning round $t$, only the current task $\mathcal{D}^t=\{\mathcal{X}^t,\mathcal{Y}^t\}$ of paired images and masks is accessible, while past tasks $\{\mathcal{D}^i\}_{i=1}^{t-1}$ are unavailable.
Our objective is to continually train $f_{\theta}$ on each incoming task without sacrificing performance on previous tasks.

\label{sec:c2gr}
A feasible way to address forgetting in TIL is data-centric generative replay, which uses generative models to recover past-task data.
However, existing replay-based methods only synthesize images and employ asynchronous optimization between the generator and segmentor \cite{gr,likangsrdsfw}.
These methods face two limitations: 1) image-only replay is susceptible to the current-task objective by adopting error-prone masks under concurrent shifts; and 2) asynchronous optimization fails to guarantee segmentation-oriented generation, introducing generation errors that compromise knowledge retention.

To address these limitations, we propose Coupled Comprehensive Generative Replay (C$^2$GR), as shown in Fig.~\ref{fig:overall}.
C$^2$GR synthesizes image-mask pairs for previous tasks to mitigate forgetting, which requires preserving image-mask correspondence and coupling the optimization of generator and segmentor for segmentation-oriented generation.
To preserve the correspondence, we design Bayesian Joint Diffusion (BJD) that formulates the correspondence as a conditional denoising optimization objective.
To couple the optimization, we introduce Relation-aware Unified Prompt Synchronization (RUPS) that injects a shared task-relation-aware prompt into both the generator and segmentor for simultaneously updating the two models.
The following sections elaborate on each component.

\subsection{Bayesian Joint Diffusion (BJD)}
\label{sec:BIMJD}
A naive approach to jointly synthesize images and masks is to regard them as a single variable and model its distribution with diffusion models \cite{ho2020denoising}.
However, the image-mask correspondence is often disrupted during this naive joint diffusion process \cite{medsegfactory}.
We therefore revisit the naive process and propose a Bayesian joint diffusion that formulates image-mask correspondence as conditional distributions and optimizes them via conditional denoising.

\subsubsection{Naive Joint Diffusion  (NJD) process}
This naive process models the joint distribution of image-mask pairs via forward and reverse diffusion processes \cite{medsegfactory}.
It needs to be performed at each previous learning round to simulate past-task data $\mathcal{D}^{i\in[1:t-1]}$.
Given $(x_0,y_0)\!\sim\!\mathcal{D}^i$, the forward process progressively adds Gaussian noise over $K$ steps to produce noisy pairs $\{(x_k,y_k)\}_{k=1}^K$ for training a denoising network $\epsilon_\theta$.
The reverse process then employs $\epsilon_\theta$ to iteratively recover $(x_0,y_0)$ from random Gaussian noise.
The denoising network $\epsilon_\theta$ is optimized via Maximum Likelihood Estimation (MLE) on the image-mask joint distribution, i.e., $\max \log p(x,y)$.
This objective can be simplified by joint denoising score matching \cite{bao2023one} under an assumption that images and masks are nearly independent \cite{medsegfactory}, as noise added in the forward process corrupts their correlation.
At step $k$, Gaussian noise $\epsilon_x,\epsilon_y\!\sim\!\mathcal{N}(0,\boldsymbol{\mathrm{I}})$ is added to $x_0$ and $y_0$, yielding noisy samples $x_k = \sqrt{\bar{\alpha}_k}\,x_0 + \sqrt{1-\bar{\alpha}_k}\,\epsilon_x$ and $y_k = \sqrt{\bar{\alpha}_k}\,y_0 + \sqrt{1-\bar{\alpha}_k}\,\epsilon_y$ with noise level $\bar{\alpha}_k$.
The denoising network $\epsilon_\theta$ is trained to jointly predict the noise via the simplified loss:
\begin{equation} L_{\text{NJ}}^{\mathcal{D}^i} = {\mathbb{E}}_{k,\epsilon_x,\epsilon_y,(x_0,y_0)}\Big [ \Vert [{\epsilon}_{x},{\epsilon}_{y}] - {\epsilon}_\theta ([{x}_k, {y}_k], k) \Vert^2_2 \Big ],
\label{eq:L_NJ}
\end{equation}
where $[,]$ denotes concatenation along the channel dimension.

However, simultaneously adding $\epsilon_x$ and $\epsilon_y$ to the image-mask pair $(x_0,y_0)$ easily distorts the image-mask correspondence, especially as the noise level gradually grows in the forward process.
This could cause misaligned image appearance and segmentation objective, thereby hampering the synthesis of realistic anatomical structures.

\subsubsection{Bayesian Joint Diffusion (BJD) process}
To address this issue, we introduce a Bayesian joint diffusion process that employs two complementary conditional distributions to explicitly model image-mask correspondence.
By Bayes' theorem, the MLE objective can be reformulated as:
\begin{equation}
\label{eq:bj-objective}
\begin{aligned}
\max \log p(x,y)
&= \max \Bigl[\log p(x)p(y) + \\
&\phantom{= \max \Bigl[}\log p(x| y) + \log p(y| x)\Bigr],
\end{aligned}
\end{equation}
where $\max \log\!\big(p(x)p(y)\big)$ can be approximately regarded as the optimization objective of the NJD process under the independence assumption \cite{medsegfactory}, particularly when the image-mask correspondence is distorted.
Therefore, the objective of the NJD process neglects the conditional terms $\max \log p(x|y)$ and $\max \log p(y|x)$, which capture the intrinsic correspondence between images and masks.
We then modify the training loss of Eq.~\ref{eq:L_NJ} to recover this correspondence.
Specifically, $\max \log\!\big(p(x)p(y)\big)$ can be converted into the simplified loss in Eq.~\ref{eq:L_NJ} due to the independence assumption.
The conditional objectives $\max \log p(x|y)$ and $\max \log p(y|x)$ are simplified via conditional denoising score matching \cite{ho2022classifier}, where the noise-free mask $y_0$ and image $x_0$ alternately serve as references rather than adding noise to both simultaneously.
\begin{align}
L_{\text{BJ}}^{\mathcal{D}^i}
= {\mathbb{E}}_{k,\epsilon_x,\epsilon_y,(x_0,y_0)}\Bigl[
& \Vert [{\epsilon}_{x},{\epsilon}_{y}] - {\epsilon}_\theta ([{x}_k, {y}_k], k) \Vert^2_2 \nonumber\\
 + & \Vert [{\epsilon}_{x},{0}] - {\epsilon}_\theta ([{x}_k, {y}_0], k) \Vert^2_2 \nonumber\\
 + & \Vert [{0},{\epsilon}_{y}] - {\epsilon}_\theta ([{x}_0, {y}_k], k) \Vert^2_2
\Bigr].
\label{eq:L_BJD}
\end{align}
Intuitively, $\epsilon_\theta([x_k,y_0],k)$ restores image appearance aligned with the clean mask, while $\epsilon_\theta([x_0,y_k],k)$ recovers the segmentation mask consistent with the clean image.
We use the same reverse process as in the NJD process to generate image-mask pairs from random Gaussian noise.

\subsection{Relation-aware Unified Prompt Synchronization (RUPS)}
\label{sec:rupg}
Existing generative replay methods \cite{gr,likangsrdsfw} typically optimize the generator and segmentor independently.
Directly adopting this strategy may not guarantee segmentation-oriented generation to effectively preserve the segmentor knowledge.
To couple the optimization of the generator and the segmentor, we propose a Relation-aware Unified Prompt Synchronization (RUPS) method.
The proposed RUPS formulates a unified prompt via a relation-aware prompt graph to simultaneously modulate the generator and segmentor, thereby synchronizing the optimization of both models.

\subsubsection{Relation-aware Prompt Graph}
A segmentation task is described by its imaging modality, anatomical region, and segmentation objective \cite{liu2023clip}.
For each task, to capture the complex relation along these attributes with other tasks, we design a relation-aware prompt graph that merges attribute-specific relation into a predefined textual prompt \cite{radford2021learning}.
Specifically, we use a textual prompt template ``A [I] image of [A] for [SO]'' to describe each task with three attributes: imaging modality $\text{I}_i$, anatomical region $\text{A}_i$, and segmentation objective $\text{SO}_i$.
To capture attribute-specific relation, we define six subgraphs $\{\Phi_S\}_{S\in\mathcal{M}}$ indexed by attribute subsets
$\mathcal{M}{=}\big\{\{\text{I}\},\{\text{A}\},\{\text{SO}\},\{\text{I},\text{A}\},\{\text{I},\text{SO}\},\{\text{A},\text{SO}\}\big\}$.
Each subgraph captures task relation along specific attributes over text prompts of all tasks stored in a prompt pool.
For instance, $\Phi_{\{\text{I}\}}$ captures relation on modality, while $\Phi_{\{\text{I},\text{A}\}}$ captures relation on both modality and anatomical region.
For task $i$, its prompt is encoded into embedding $e_i$ via a frozen CLIP text encoder \cite{radford2021learning}.
The $e_i$ is projected into a graph node $n_i$ via a learnable projection $W_E$ shared across all subgraphs, i.e., $n_i {=} W_E \cdot e_i$.
In each subgraph $\Phi_S$, a linear layer $V_{\Phi_S}$ computes attention scores $a^{\Phi_S}_{i_1,i_2}$ between nodes $n_{i_1}$ and $n_{i_2}$ of task $i_1$ and $i_2$:
\begin{equation}
    \label{eq:attention_coeff}
   a^{\Phi_S}_{i_1,i_2} = \eta \cdot \frac{\exp (V_{\Phi_S} (n_{i_1}, n_{i_2}))}{\sum_{i\in[1:t]} \exp (V_{\Phi_S} (n_{i_1}, n_{i}))} + (1-\eta) \cdot s^{\Phi_S}_{i_1,i_2},
\end{equation}
where $s^{\Phi_S}_{i_1,i_2}$ denotes the cosine similarity between the attribute-related components of $e_{i_1}$ and $e_{i_2}$ under $\Phi_S$, and $\eta{\in}[0,1]$ is a learnable weight balancing graph attention and $s^{\Phi_S}_{i_1,i_2}$.
The embedding $z^{\Phi_S}_i$ of task $i$ in subgraph $\Phi_S$ is then obtained as $z^{\Phi_S}_i {\,=\,} \sigma ({\sum}_{i_t \in [1:t]} a^{\Phi_S}_{i,i_t} \cdot n_{i_t})$, where $\sigma$ denotes a LeakyReLU function.
Then, we evaluate the importance of each subgraph embedding via a score $w_{\Phi_S}$ and aggregate the embeddings to obtain the relation-aware prompt $\zeta_i$ of task $i$:
\begin{align}
w_{\Phi_S} &= \frac{1}{t}{\sum}^t_{i=1} q^{\top}\cdot\tanh\!\left(W_q\, z^{\Phi_S}_i + b_q\right), \nonumber \\
\zeta_i &= {\sum}_{S\in\mathcal{M}} \frac{\exp\!\left(w_{\Phi_S}\right)}{{\sum}_{S'\in \mathcal{M}}\exp\!\left(w_{\Phi_{S'}}\right)} \cdot z^{\Phi_S}_i,
\label{eq:prompt_aggregation}
\end{align}
where $W_q$, $b_q$, and $q$ are learnable parameters.

\subsubsection{Unified Prompt Modulation}
After obtaining the relation-aware task prompt $\zeta_i$, it is simultaneously injected into both the segmentation model and the denoising network through a cross-attention mechanism \cite{sd}.
Specifically, taking the segmentation model $f_\theta$ as an example, in each layer $l$ of the $f_\theta$, the intermediate model feature $h_l$ is combined with the task prompt $\zeta_i$ through cross-attention:
\begin{equation}
\tau_l = h_l + \phi\!\left(\frac{(Q_l h_l)(K_l \zeta_i)^\top}{\sqrt{d_l}}\right)(V_l \zeta_i),
\end{equation}
where $\tau_l$ denotes the output of layer $l$, $\phi(\cdot)$ is the softmax function, $Q_l$, $K_l$, and $V_l$ are learnable query, key, and value of cross-attention, and $d_l$ is the dimension of cross-attention module.
The task prompt $\zeta_i$ is also injected into the denoising network $\epsilon_\theta$ in the same way.
By simultaneously modulating both the segmentation model and the denoising network, the two models can interact through the shared task prompt.
Consequently, after the two models are jointly optimized, the shared prompt can guide the denoising network to generate segmentation-oriented image-mask pairs for each task.

\subsection{Overall Training Pipeline}
\label{sec:training}
At each learning round $t$, the overall training pipeline consists of three stages.
First, we leverage the prompts of previous tasks obtained in previous rounds to modulate the denoising network to synthesize image-mask pairs, forming the replay datasets $\mathcal{R}^{i\in[1:t-1]}$.
Second, we use the relation-aware prompt graph to generate the prompt $\zeta_i$ for the current task $i$.
Third, the segmentation model $f_\theta$ and the denoising network $\epsilon_\theta$ are jointly optimized on both the current task $\mathcal{D}^t$ and the replay datasets $\mathcal{R}^{i\in[1:t-1]}$.
Specifically, we use the task prompt $\zeta_{j{\in}[1,t]}$ to simultaneously modulate both $f_\theta$ and $\epsilon_\theta$, forming $f^{\zeta_j}_\theta$ and $\epsilon^{\zeta_j}_\theta$, as introduced in the RUPS method.
The segmentor $f_\theta$ is trained with a segmentation loss $L_{\text{seg}} = L_{\text{dice}} + L_{\text{ce}}$, where $L_{\text{dice}}$ is a Dice loss and $L_{\text{ce}}$ is a cross-entropy loss.
We term this process Memory Evoking:
\begin{equation}
\label{eq:me_overview}
L_{\text{ME}} = L^{\mathcal{D}^t}_{\text{seg}}(f^{\zeta_t}_\theta) + \frac{1}{t-1}{\sum}_{i=1}^{t-1} L^{\mathcal{R}^i}_{\text{seg}}(f^{\zeta_i}_\theta).
\end{equation}
Meanwhile, the denoising network $\epsilon_\theta$ is also updated on $\mathcal{D}^t$ and $\mathcal{R}^{i\in[1:t-1]}$ to simulate the current task for subsequent replays. We term this process Memory Updating:
\begin{equation}
\label{eq:mu_overview}
L_{\text{MU}} = L_{\text{BJ}}^{\mathcal{D}^t}(\epsilon^{\zeta_t}_\theta) + \frac{1}{t-1}{\sum}_{i=1}^{t-1} L_{\text{BJ}}^{\mathcal{R}^i}(\epsilon^{\zeta_i}_\theta),
\end{equation}
where $L_{\text{BJ}}$ is the loss of the BJD process as shown in Eq.~\ref{eq:L_BJD}.
Then the $f_\theta$ and $\epsilon_\theta$ are jointly updated via the following loss:
\begin{equation}
\label{eq:overall}
L = L_{\text{ME}} + L_{\text{MU}}.
\end{equation}

% !TEX root = ../main.tex
\section{Experiments}
\label{sec:exp}

\subsection{Datasets and Experiment Settings}
\subsubsection{Datasets}
\label{sec:dataset}

\begin{figure}[!htbp]
    \centering
\includegraphics[width=0.79\columnwidth]{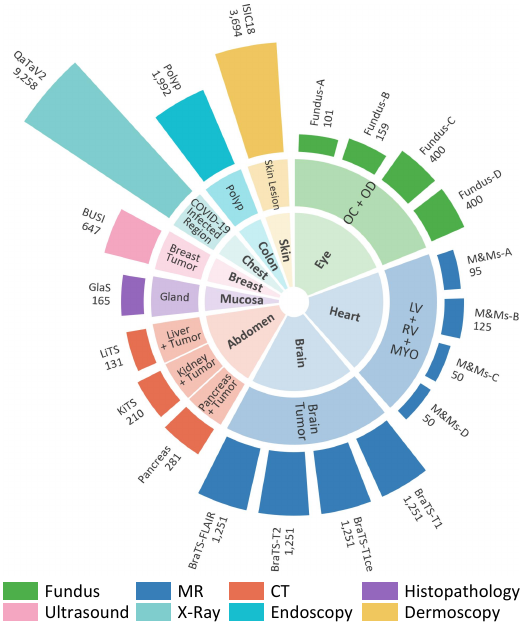}
    \caption{Segmentation tasks involved during training. OC and OD denote optic cup and disc. LV, RV, and MYO denote left ventricle, right ventricle, and myocardium. M\&Ms-A, B, C, D are vendors A to D of M\&Ms. BraTS-T1, T1ce, T2, and FLAIR are MR sequences of BraTS.}
    \label{fig:dataset_distribution}
\end{figure}

We collect a wide range of public medical segmentation datasets covering diverse modalities and anatomical regions.
A multi-site fundus dataset \cite{dofe} for optic cup and disc segmentation provides cross-site appearance shift, with four subsets from Aravind Eye Hospital (\textbf{Fundus-A}), a Nidek AFC-210 camera (\textbf{Fundus-B}), Canon CR-2 (\textbf{Fundus-C}), and Zeiss Visucam 500 (\textbf{Fundus-D}).
\textbf{M\&Ms} \cite{mnms} cine MR from four vendors (Siemens, Philips, GE, and Canon) targets cardiac segmentation, and \textbf{BraTS} \cite{baid2021rsna} multi-sequence MR (T1, T1ce, T2, FLAIR) targets brain tumor segmentation.
\textbf{Pancreas} \cite{antonelli2022medical}, \textbf{KiTS} \cite{kits}, and \textbf{LiTS} \cite{antonelli2022medical} cover pancreas and pancreatic tumor, kidney and kidney tumor, and liver and liver tumor segmentation in abdominal CT, respectively.
\textbf{GlaS} \cite{glas} targets colon gland segmentation in histopathology, \textbf{BUSI} \cite{busi} targets ultrasound breast tumor, \textbf{QaTaV2} \cite{qatav2} targets COVID-19 region in chest X-ray, \textbf{Polyp} \cite{polypdata} targets endoscopy polyp segmentation, and \textbf{ISIC18} \cite{codella2019skin} targets dermoscopic skin lesion segmentation. Details are shown in Fig.~\ref{fig:dataset_distribution}.
For each dataset, we follow the official train/test split and sample 15\% for validation. For datasets without an official split, we use a 60\%/15\%/25\% split for training, validation, and testing.
Each dataset is preprocessed via its official pipeline including center-cropping, resizing, and normalization, and all images and axial slices of 3D volumes are resized to $256\times256$.

\subsubsection{Experimental Setting and Evaluation Metrics}
We arrange these datasets into a sequential task stream from Fundus-A to ISIC18 following Fig.~\ref{fig:dataset_distribution}. 
The segmentation model is updated sequentially and past tasks become inaccessible once a new task arrives.
We evaluate segmentation performance using the Dice Similarity Coefficient (DSC), and report the average DSC over foreground categories for datasets with multiple foreground categories.
We further build a train-test matrix $D \in \mathbb{R}^{T\times T}$, with $D_{i,j}$ the DSC on task $j$ after training up to task $i$, and report the Overall, Backward Transfer (BWT), and Transfer Learning (TL):
\begin{equation}
\begin{aligned}
\mathrm{Overall} &= \frac{\sum_{j=1}^{T}D_{T,j}}{T}, \\
\mathrm{BWT} &= \frac{2\sum_{i=2}^{T}\sum_{j=1}^{i-1}\left(1-\left|\min\!\left(D_{i,j}-D_{j,j},0\right)\right|\right)}{T(T-1)}, \\
\mathrm{TL} &= \frac{\sum_{i=1}^{T}D_{i,i}}{T}.
\end{aligned}
\end{equation}
Overall is the mean DSC over all tasks after final-task training. BWT measures stability against catastrophic forgetting. TL reflects plasticity when learning each task.

\subsection{Implementation Details}
We adopted the UNet \cite{unet} as the backbone for both the segmentor with channel numbers of 32, 64, 128, 256, 512, 512, 512 and the denoising network with channel numbers of 128, 128, 256, 256, 512, 512, both equipped with cross-attention layers of dimension 768.
We set the forward diffusion step to $K{=}1000$. The dimensions of the graph node and the learnable vector $q$ were set to 512 and 256, respectively.
For tasks sharing the same modality but from different sites or devices, we incorporated the site or device information into the task prompt.
During training, the reverse process was set to 200 steps to generate the same number of image-mask pairs for each past task as in its training set.
Both models were optimized by Adam optimizer with a learning rate of $1\!\times\!10^{-4}$ and a batch size of 48 for 300 epochs per round.
All experiments were conducted on two NVIDIA A100 GPUs.

\subsection{Comparison with Existing Methods}

\subsubsection{Compared Methods}
We compare our Coupled Comprehensive Generative Replay (\textbf{C$^2$GR}) against existing methods:
1) Baselines: \textbf{JointTrain}, which gathers all task data for joint training; \textbf{FineTune}, which sequentially finetunes the segmentation model using only current task data; and \textbf{IndividualTrain}, which stores task-specific models and selects the corresponding one during inference;
2) DIL schemes: \textbf{EWC} \cite{kirkpatrick2017overcoming}, which uses appearance regularization to preserve learned knowledge; \textbf{GR} \cite{gr}, which performs image-only replay without generating corresponding masks; \textbf{SR-DSFW} \cite{likangsrdsfw}, which replays images according to current task masks; and \textbf{TED} \cite{ted}, using tri-enhanced appearance regularization at the distillation, transfer, and fusion levels;
and 3) CIL schemes: \textbf{PLOP} \cite{plop}, which uses feature-level knowledge distillation; \textbf{MEIL} \cite{meil}, which transfers class prototypes; \textbf{CoNuSeg} \cite{conuseg}, which applies distillation to both model features and class prototypes; and \textbf{MedSeqFT} \cite{ye2025medseqft}, which uses distillation and memory replay for continual learning on foundation models.
We also include \textbf{CGR} \cite{cgr}, our previous version that performs comprehensive generative replay for image-mask pairs with asynchronous optimization of the generator and segmentor.

\subsubsection{Experimental Results}

% !TEX root = ../main.tex
\begin{table*}[!t]
    \centering
    \caption{Performance on the 20 sequential tasks from Fundus-A to ISIC18. \textbf{Bold} marks the best among compared incremental learning methods. $\dagger$ denotes the previous conference version of our method.}
    \renewcommand{\arraystretch}{1.1}
    \setlength{\tabcolsep}{2pt}
    \newcommand{\taskhead}[2]{\begin{tabular}[c]{@{}c@{}}#1\\[-0.85ex]#2\end{tabular}}
    \resizebox{\linewidth}{!}{
    \begin{tabular}{l|ccccccccccccccccccc|c|c|c|c}
        \hline \hline
         \multicolumn{1}{c|}{} & \multicolumn{19}{c|}{Previous} & \multicolumn{1}{c|}{Incoming} & \multirow{2}{*}{Overall} & \multirow{2}{*}{BWT} & \multirow{2}{*}{TL} \\
        \cline{2-21}
          & \taskhead{Fundus}{A} & \taskhead{Fundus}{B} & \taskhead{Fundus}{C} & \taskhead{Fundus}{D} & \taskhead{M\&Ms}{A} & \taskhead{M\&Ms}{B} & \taskhead{M\&Ms}{C} & \taskhead{M\&Ms}{D} & \taskhead{BraTS}{T1} & \taskhead{BraTS}{T1ce} & \taskhead{BraTS}{T2} & \taskhead{BraTS}{FLAIR} & Pancreas & KiTS & LiTS & GlaS & BUSI & QaTaV2 & Polyp & ISIC18 &  &  &  \\
        \hline \hline

        IndividualTrain & 82.87 & 84.39 & 89.36 & 90.72 & 90.86 & 89.94 & 90.27 & 92.63 & 75.62 & 80.73 & 89.42 & 89.47 & 75.17 & 90.95 & 79.40 & 89.32 & 78.07 & 78.75 & 90.36 & 87.77 & 85.80 & 100.00 & 85.80 \\
        JointTrain & 89.59 & 86.29 & 89.47 & 91.14 & 90.83 & 91.30 & 91.17 & 92.27 & 77.67 & 81.18 & 89.17 & 89.98 & 72.07 & 89.15 & 81.23 & 88.88 & 76.34 & 77.82 & 88.13 & 87.59 & 86.06 & 100.00 & 86.06 \\
        FineTune & 27.82 & 17.79 & 20.85 & 11.73 & 6.76 & 6.75 & 7.93 & 3.67 & 17.53 & 18.12 & 19.66 & 16.76 & 31.73 & 35.63 & 6.73 & 47.84 & 31.74 & 39.58 & 48.01 & 88.02 & 25.23 & 30.98 & 84.50 \\
        \hline

        EWC \cite{kirkpatrick2017overcoming} & 30.99 & 18.07 & 19.27 & 12.19 & 8.62 & 7.38 & 8.57 & 5.69 & 17.63 & 16.19 & 28.60 & 22.72 & 32.44 & 36.56 & 17.82 & 41.38 & 34.08 & 44.26 & 46.59 & 85.79 & 26.74 & 36.46 & 81.07 \\
        GR \cite{gr} & 81.72 & 77.01 & 84.16 & 86.11 & 86.54 & 87.52 & 86.53 & 88.60 & 69.39 & 69.85 & 82.51 & 84.86 & 51.31 & 71.77 & 69.86 & 63.76 & 49.41 & 65.46 & 66.45 & 87.89 & 75.54 & 93.78 & 82.52 \\
        SR-DSFW \cite{likangsrdsfw} & 28.40 & 18.13 & 20.00 & 20.19 & 23.51 & 20.85 & 23.66 & 19.61 & 58.75 & 72.04 & 78.67 & 85.15 & 41.15 & 34.18 & 12.00 & 43.46 & 58.38 & 37.41 & 70.62 & 87.96 & 42.71 & 43.66 & 84.37 \\
        TED \cite{ted} & 19.55 & 7.51 & 8.12 & 4.55 & 11.95 & 9.73 & 12.53 & 7.72 & 17.59 & 18.68 & 25.68 & 17.22 & 32.34 & 39.16 & 3.88 & 32.25 & 34.62 & 41.53 & 43.62 & \textbf{87.96} & 23.81 & 33.13 & 83.71 \\
        \hline

        PLOP \cite{plop} & 75.13 & 44.80 & 69.84 & 51.52 & 2.73 & 10.99 & 12.09 & 1.79 & 35.09 & 29.89 & 50.26 & 75.27 & 34.15 & 43.17 & 60.59 & 66.46 & 52.02 & 59.51 & 62.81 & 86.80 & 46.25 & 68.03 & 82.46 \\
        MEIL \cite{meil} & 75.18 & 61.70 & 71.52 & 75.65 & 27.05 & 27.05 & 19.17 & 8.11 & 53.01 & 57.41 & 66.18 & 80.43 & 41.22 & 56.26 & 64.06 & 76.47 & 60.29 & 64.76 & 66.33 & 86.72 & 56.93 & 79.96 & 82.94 \\
        CoNuSeg \cite{conuseg} & 75.96 & 65.97 & 79.09 & 80.30 & 56.60 & 55.17 & 48.59 & 45.29 & 61.90 & 69.35 & 52.98 & 77.99 & 35.94 & 47.30 & 40.08 & 38.87 & 44.33 & 64.31 & 60.74 & 84.39 & 59.26 & 84.97 & 80.11 \\
        MedSeqFT \cite{ye2025medseqft} & 82.62 & 73.15 & 85.59 & 85.28 & 87.27 & \textbf{89.53} & 88.95 & \textbf{90.08} & 71.21 & 75.79 & 85.26 & 86.98 & 55.17 & 78.68 & 75.31 & 50.46 & 50.29 & 69.61 & 73.95 & 87.33 & 77.13 & 95.18 & 84.57 \\
        \hline

        CGR$^\dagger$ \cite{cgr} & \textbf{85.08} & 77.77 & 87.18 & 87.22 & 85.63 & 88.42 & 88.42 & 88.42 & 75.11 & 80.81 & 87.46 & 88.70 & 69.08 & \textbf{84.80} & 77.41 & 84.13 & 70.38 & 70.41 & 73.54 & 87.92 & 81.89 & 96.14 & 84.16 \\
        \textbf{C$^2$GR (ours)} & 84.58 & \textbf{84.25} & \textbf{87.41} & \textbf{88.41} & \textbf{87.72} & 88.64 & \textbf{89.40} & 89.83 & \textbf{77.94} & \textbf{82.79} & \textbf{87.60} & \textbf{89.93} & \textbf{73.84} & 84.64 & \textbf{77.98} & \textbf{84.68} & \textbf{71.39} & \textbf{74.67} & \textbf{79.18} & 87.53 & \textbf{83.62} & \textbf{98.16} & \textbf{85.24} \\
        \hline \hline
    \end{tabular}
    }
    \label{tab:comparison}
\end{table*}

\begin{figure*}[!htbp]
    \centerline{\includegraphics[width=0.9\textwidth]{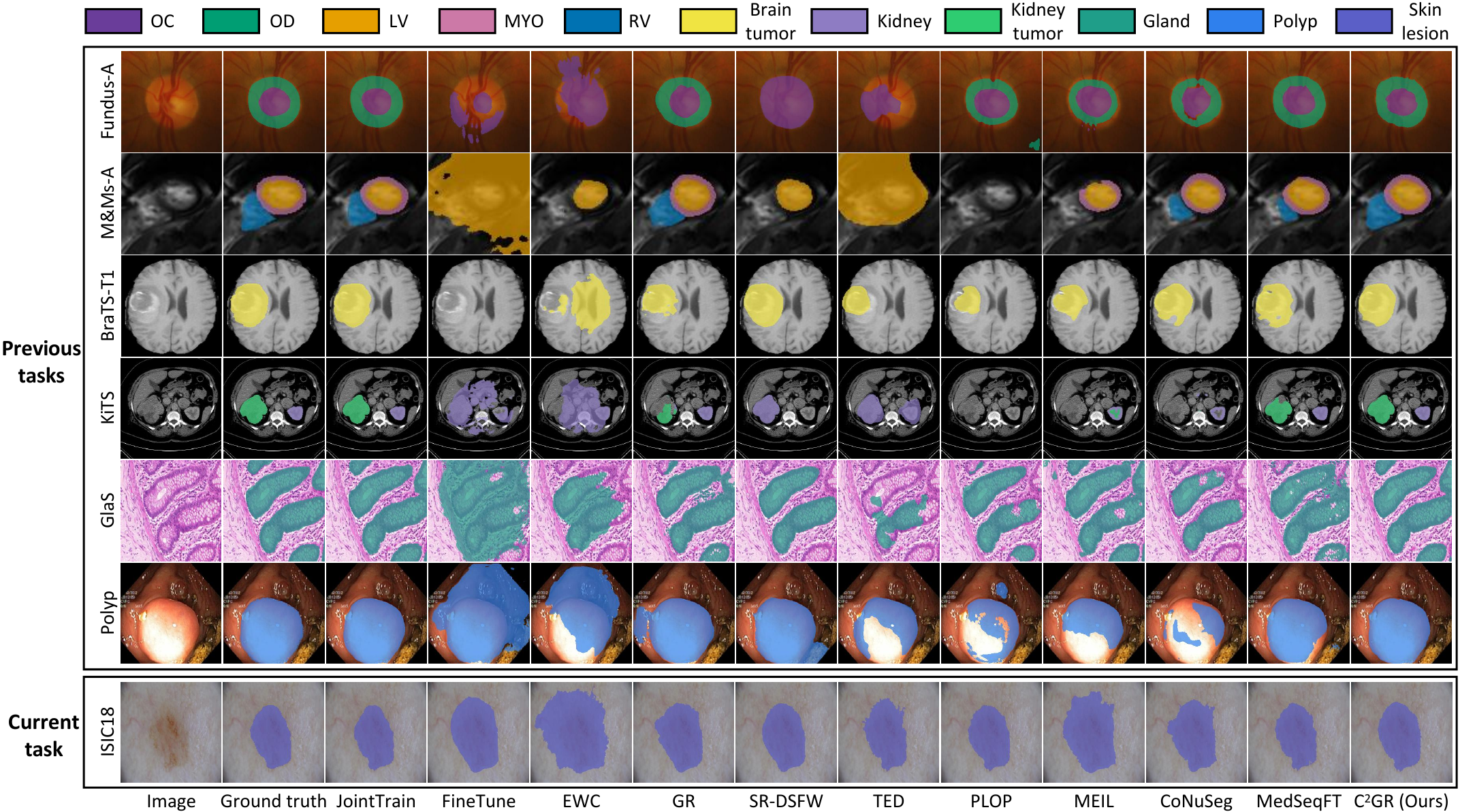}}
    \caption{Qualitative comparison on 20 sequential tasks (Fundus-A to ISIC18). Different colors mark different objectives. Abbreviations follow Fig.~\ref{fig:dataset_distribution}.}
    \label{fig:seg}
\end{figure*}

Quantitative results on the 20 sequential tasks are reported in Table~\ref{tab:comparison}.
Most methods perform comparably on the incoming task, but FineTune collapses on previous tasks under concurrent appearance and objective shifts. JointTrain slightly outperforms IndividualTrain by exploiting task correlations within a single universal model.
Among DIL methods, appearance regularization methods (e.g., EWC and TED) fail to preserve knowledge related to the segmentation objective, resulting in severe degradation compared to JointTrain.
Image-only replay methods (e.g., GR and SR-DSFW) also struggle to maintain the performance of segmentor on past tasks, as they reuse error-prone masks dominated by the current task objective under objective shift.
Among CIL methods, objective-specific distillation methods (e.g., PLOP, MEIL, and CoNuSeg) also degrade significantly, as the appearance shift undermines feature distillation and prototype transfer.
MedSeqFT does not exhibit significant degradation by storing raw past-task data, which is often restricted by data-sharing constraints.
In contrast, C$^2$GR effectively mitigates forgetting by simultaneously generating past-task image-mask pairs, surpassing the second-best method MedSeqFT by 6.49\% in Overall DSC and 2.98\% in BWT.
Meanwhile, our method approaches the plasticity of JointTrain, as evidenced by the comparable TL.
Furthermore, the proposed method surpasses our conference version CGR by 1.73\% Overall DSC, as C$^2$GR couples the optimization of generator and segmentor to synthesize segmentation-oriented image-mask pairs.
Fig.~\ref{fig:seg} qualitatively shows that C$^2$GR yields predictions closest to ground truth and most consistent with JointTrain.

\subsection{Detail Analysis}
\subsubsection{Effectiveness of Each Module}
\label{sec:ablation_module}

\begin{figure}[!htbp]
    \centering
    \includegraphics[width=0.7\linewidth]{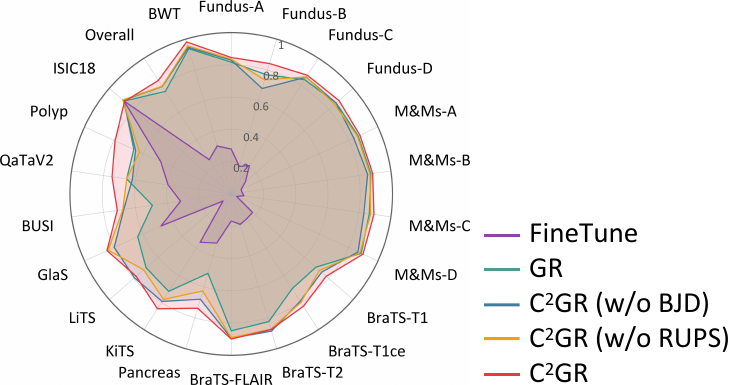}
    \caption{Ablation study across 20 tasks examining the contributions of BJD and RUPS in C$^2$GR.}
    \label{fig:ablation_module}
\end{figure}

To validate the effectiveness of each module, we compare C$^2$GR with two variants: 1) \textbf{C$^2$GR (w/o BJD)}, which uses NJD to generate past-task data; and 2) \textbf{C$^2$GR (w/o RUPS)}, which directly adopts CLIP-based task prompts for the segmentor and generator without coupling their optimization processes.
As shown in Fig.~\ref{fig:ablation_module}, compared with FineTune and GR, all variants achieve improvements, indicating that replaying image-mask pairs for previous tasks mitigates forgetting.
However, removing either BJD or RUPS decreases the performance on previous tasks compared to C$^2$GR.
Such degradation stems from the absence of BJD for preserving image-mask correspondence or RUPS for coupling the optimization of generator and segmentor to reinforce segmentation-oriented generation via relation prior.

\subsubsection{Quality Evaluation of the Replayed Image-Mask Pairs}
\label{sec:gen_quality}
\begin{figure*}[!t]
    \centering
    \includegraphics[width=0.9\textwidth]{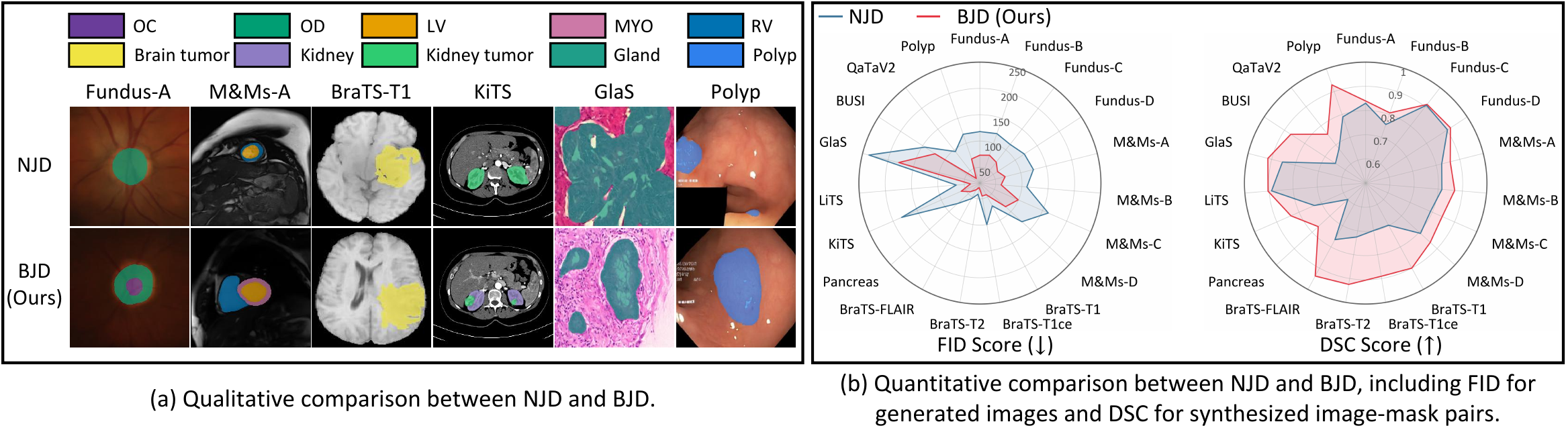}
    \caption{Qualitative and quantitative comparison between NJD and BJD. Abbreviations follow Fig.~\ref{fig:dataset_distribution}.}
    \label{fig:gen_sample}
\end{figure*}

We visualize synthesized image-mask pairs of previous tasks by NJD and BJD after training on the final task.
As shown in Fig.~\ref{fig:gen_sample}(a), NJD produces image-mask pairs with weaker correspondence, where the masks exhibit semantic errors and do not correspond to the image content.
In contrast, BJD yields semantically correct masks that faithfully reflect the images.
We further quantify image generation quality using Fr\'{e}chet Inception Distance (FID) between the synthesized and real images, and assess image-mask correspondence via the DSC between the generated masks and the outputs from the JointTrain model.
As illustrated in Fig.~\ref{fig:gen_sample}(b), BJD consistently achieves lower FID and higher DSC than NJD, showing that conditional denoising guided by clean images and masks improves both image realism and image-mask correspondence.

\subsubsection{Analysis of the Relation-aware Unified Task Prompt}
\label{sec:rupl_analysis}

\begin{figure}[!htbp]
    \centering
    \includegraphics[width=\columnwidth]{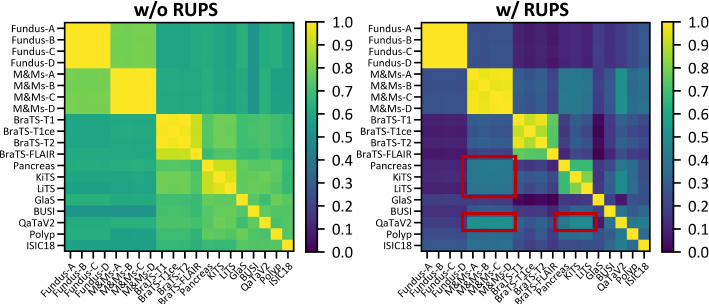}
    \caption{Cosine similarity of task prompts with and without RUPS.}
    \label{fig:htrg_vis}
\end{figure}

\begin{figure}[!htbp]
    \centering
    \includegraphics[width=\columnwidth]{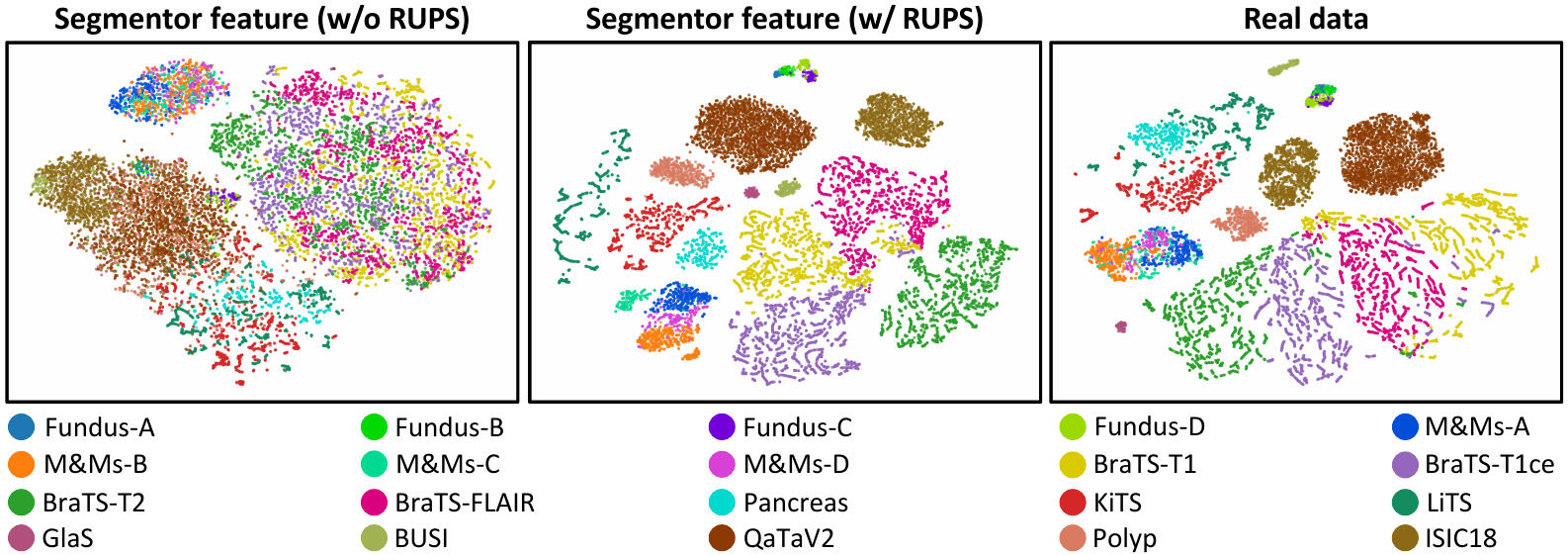}
    \caption{t-SNE visualization of segmentor features with and without RUPS, alongside real data for each task.}
    \label{fig:tsne_vis}
\end{figure}

Fig.~\ref{fig:htrg_vis} compares the cosine similarity of task prompts with and without RUPS.
Without RUPS, CLIP embeddings produce uniformly high similarity scores that fail to distinguish these tasks.
In contrast, the relation-aware task prompts from RUPS capture meaningful inter-task relation.
For instance, among tasks sharing the same anatomical region and segmentation objective yet differing in imaging protocol, the corresponding prompts exhibit higher similarity, as reflected in the diagonal blocks (e.g., the four fundus tasks).
Beyond the diagonal blocks, the relation-aware prompts also capture plausible task relation, as shown in red boxes.
Specifically, the prompts of QaTaV2 and M\&Ms exhibit high similarity owing to their similar anatomical regions (chest and heart).
Abdominal CT task prompts also exhibit high similarity with QaTaV2 and M\&Ms, given their spatially adjacent anatomical regions (abdomen, chest, and heart) and similar modalities (CT, X-ray, and MR).
We further visualize the t-SNE of segmentor features in Fig.~\ref{fig:tsne_vis}.
With RUPS, the segmentor produces more discriminative features across tasks while preserving task relation similar to those in real data distributions.
These results show that the unified task prompt facilitates generating segmentation-oriented image-mask pairs during coupled optimization, bringing the feature distribution of the segmentor closer to that of real data and thereby enhancing knowledge retention for previous tasks.

\subsubsection{Adaptation to Downstream Tasks}
\label{sec:downstream}

To validate the adaptability of models trained by different methods to unseen tasks, we select five downstream datasets covering diverse modalities and objectives (objective, modality): \textbf{DRIVE} \cite{drive} (retinal vessels, fundus), \textbf{GOALS} \cite{goals} (retinal layers, OCT), \textbf{SegPC} \cite{segpc} (cytoplasm and nucleus, microscopy), \textbf{DDTI} \cite{ddti} (thyroid nodules, ultrasound), and \textbf{PROMISE12} \cite{promise} (prostate, MR). All are preprocessed following Sec.~\ref{sec:dataset}.
We compare C$^2$GR with training from Scratch, two baselines (JointTrain, FineTune), the top two DIL methods (GR, SR-DSFW), and the top two CIL methods (CoNuSeg, MedSeqFT). 
Each task is fine-tuned for 60 epochs at a learning rate of $10^{-4}$ for fast adaptation. 
Fine-tuning is initialized from the model trained through all upstream tasks.
As shown in Table~\ref{tab:downstream}, all methods outperform Scratch, indicating upstream knowledge is transferred to downstream tasks.
However, existing methods yield limited performance due to knowledge forgetting on upstream tasks.
In contrast, C$^2$GR achieves the best performance among all methods and approaches JointTrain by preserving more complete knowledge.

% !TEX root = ../main.tex
\begin{table}[!htbp]
    \centering
    \caption{DSC (\%) on 5 downstream tasks. \textbf{Bold} marks the best among compared incremental learning methods.}
    \renewcommand{\arraystretch}{1.0}
    \setlength{\tabcolsep}{2pt}
    \resizebox{0.95\linewidth}{!}{
    \begin{tabular}{l|ccccc|c}
        \hline \hline
         & DRIVE & GOALS & SegPC & DDTI & PROMISE12 & Overall \\
        \hline \hline

        Scratch & 70.73 & 87.81 & 73.21 & 74.49 & 72.08 & 75.66 \\
        \hline
        JointTrain & 73.08 & 91.11 & 77.40 & 83.35 & 85.28 & 82.04 \\
        FineTune & 71.45 & 88.72 & 74.65 & 77.27 & 80.14 & 78.45 \\
        \hline

        GR & 72.57 & 90.26 & 75.39 & 78.68 & 81.63 & 79.71 \\
        SR-DSFW & 65.78 & 86.22 & 76.17 & 82.06 & 84.06 & 78.86 \\
        \hline

        CoNuSeg & 72.97 & \textbf{90.86} & \textbf{76.92} & 77.86 & 81.59 & 80.04 \\
        MedSeqFT & 71.93 & 89.86 & 75.66 & 79.73 & 84.43 & 80.32 \\
        \hline

        \textbf{C$^2$GR (ours)} & \textbf{74.29} & 90.76 & 76.33 & \textbf{82.25} & \textbf{85.68} & \textbf{81.86} \\
        \hline \hline
    \end{tabular}
    }
    \label{tab:downstream}
\end{table}

\subsubsection{Computational Cost}
\label{sec:compute_cost}

\begin{figure}[!htbp]
\centering
\includegraphics[width=0.8\columnwidth]{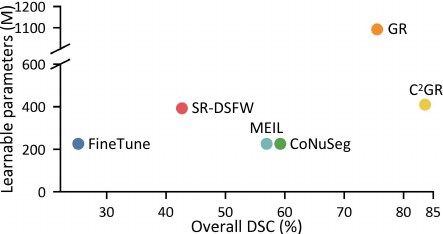}
\caption{Computational cost of different methods in terms of the number of learnable parameters with respect to their segmentation performance.}
\label{fig:compute_cost}
\end{figure}

To demonstrate the efficiency of C$^2$GR, we evaluate computational cost by measuring the number of learnable parameters.
Fig.~\ref{fig:compute_cost} presents the computational cost alongside the Overall DSC score. Although non-replay methods (e.g., MEIL and CoNuSeg) incur lower cost, they achieve poor segmentation performance, thus failing to build a universal model under concurrent shifts in image appearance and segmentation objective.
Notably, image-only replay methods (e.g., GR) exhibit limited performance even though utilizing a stronger generator (Stable Diffusion \cite{sd}) with more parameters.
In contrast, C$^2$GR achieves superior performance with comparable or lower cost.
Overall, C$^2$GR achieves high efficiency while attaining superior performance.

% !TEX root = ../main.tex
\section{Discussion}
\label{sec:discussion}

A universal model that addresses diverse segmentation tasks from different anatomical regions with varying imaging modalities and segmentation objectives has attracted increasing attention in the medical field.
Nevertheless, the prevailing paradigm for training such a universal model adheres to a static scheme that first aggregates all task data and then conducts training.
This paradigm conflicts with stringent data-sharing policies and the need to accommodate newly emerging tasks.
To bridge this gap, we propose Coupled Comprehensive Generative Replay (C$^2$GR), a task-incremental learning framework to train a universal model from sequentially arriving tasks with concurrent appearance and objective shifts.
C$^2$GR mitigates forgetting caused by such concurrent shifts by synthesizing image-mask pairs for previously learned tasks.
To achieve this, C$^2$GR preserves the image-mask correspondence of synthesized pairs and couples optimization of the segmentor and generator for segmentation-oriented generation.

On one hand, BJD promotes the generator to produce more realistic images and semantically correct masks that correspond to the images compared to NJD, as evidenced in Sec.~\ref{sec:ablation_module} and Sec.~\ref{sec:gen_quality}.
This stems from BJD formulating image-mask correspondence as conditional distributions grounded in Bayes' theorem and optimizing them via conditional denoising, whereas NJD disrupts such correspondence by adding noise to both image and mask during forward diffusion.
On the other hand, RUPS couples optimization of the generator and segmentor to guarantee segmentation-oriented replay, effectively preserving segmentor knowledge for previous tasks, as shown in Sec.~\ref{sec:ablation_module} and Sec.~\ref{sec:rupl_analysis}.
RUPS injects a relation-aware unified task prompt into both models, where the prompt encodes task-relation prior through segmentor optimization to enhance segmentation, and the segmentation-oriented prior jointly promotes segmentation-oriented synthesis through generator optimization.

Regarding computational cost, although existing non-replay methods consolidate model features or parameters without additional learnable parameters, they target settings where only appearance or objective shifts occur.
These methods degrade significantly compared to JointTrain under TIL with concurrent shifts, as shown in Table~\ref{tab:comparison}.
Replay-based methods introduce a generator to synthesize past-task images for knowledge preservation.
However, image-only replay strategies remain vulnerable to objective shift even with advanced generators at higher cost, as evidenced in Sec.~\ref{sec:compute_cost}.
Compared to these, C$^2$GR achieves superior performance at comparable or lower cost by generating paired images and masks.
Moreover, when adapting to downstream tasks under limited computational resources, the model trained with C$^2$GR achieves the best adaptation performance, as it preserves more complete upstream knowledge, as verified in Sec.~\ref{sec:downstream}.

This work focuses on building a universal segmentation model via task-incremental learning, where the key challenge is mitigating forgetting under concurrent appearance and objective shifts.
We collect diverse tasks spanning different modalities and objectives, and preprocess task data to a fixed resolution due to computational constraints.
This does not alter the nature of concurrent shifts across tasks.
However, the fixed-resolution preprocessing pipeline may not reflect the resolution needs of different modalities.
In future work, we will explore variable-resolution inputs under the TIL paradigm, e.g., employing transformer-based backbones with visual token pruning \cite{tokenpruning} to preserve key visual information while reducing computational burden, especially for high-resolution inputs.

% !TEX root = ../main.tex
\section{Conclusion}
\label{sec:conclusion}

We propose Coupled Comprehensive Generative Replay (C$^2$GR), a novel task-incremental learning framework for training a universal segmentation model under concurrent shifts in image appearance and segmentation objective.
C$^2$GR mitigates forgetting caused by the concurrent shifts by synthesizing image-mask pairs of previous tasks.
To achieve this, we propose Bayesian Joint Diffusion (BJD) to preserve the image-mask correspondence of the synthesized pairs, and introduce Relation-aware Unified Prompt Synchronization (RUPS) to couple the generator and segmentor optimization for guaranteeing segmentation-oriented generation.
Experiments show that C$^2$GR outperforms existing DIL and CIL methods in mitigating forgetting caused by the concurrent shifts.

{
    \small
    \bibliographystyle{ieeenat_fullname}
    \bibliography{IEEEabrv,reference}
}

\end{document}